# NEAR: Named Entity and Attribute Recognition of clinical concepts

Namrata Nath[*]    Sang-Heon Lee[†]

Ivan Lee[†]

**ABSTRACT**

Named Entity Recognition (NER) or the extraction of concepts from clinical text is the task of identifying entities in text and slotting them into categories such as problems, treatments, tests, clinical departments, occurrences (such as admission and discharge) and others. NER forms a critical component of processing and leveraging unstructured data from Electronic Health Records (EHR). While identifying the spans and categories of concepts is itself a challenging task, these entities could also have attributes such as negation that pivot their meanings implied to the consumers of the named entities. There has been little research dedicated to identifying the entities and their qualifying attributes together. This research hopes to contribute to the area of detecting entities and their corresponding attributes by modelling the NER task as a supervised, multi-label tagging problem with each of the attributes assigned tagging sequence labels. In this paper, we propose 3 architectures to achieve this multi-label entity tagging: BiLSTM n-CRF, BiLSTM-CRF-Smax-TF and BiLSTM n-CRF-TF. We evaluate these methods on the 2010 i2b2/VA and the i2b2 2012 shared task datasets. Our different models obtain best NER F1 scores of 0. 894 and 0.808 on the i2b2 2010/VA and i2b2 2012 respectively. The highest span based micro-averaged F1 polarity scores obtained were 0.832 and 0.836 on the i2b2 2010/VA and i2b2 2012 datasets respectively, and the highest macro-averaged F1 polarity scores obtained were 0.924 and 0.888 respectively. The modality studies conducted on i2b2 2012 dataset revealed high scores of 0.818 and 0.501 for span based micro-averaged F1 and macro-averaged F1 respectively.

[*]UniSA STEM, University of South Australia, GPO Box 2471, Adelaide SA 5001, Australia E-mail: namrata.nath@mymail.unisa.edu.au
[†]UniSA STEM, University of South Australia, Adelaide, Australia

INTRODUCTION

Clinical concept extraction is the identification of entities specific to the clinical domain such as drug names and names of clinical procedures amongst many others. Extracted entities could be used in applications such as decision support systems, prognosticating the course of medical conditions in patients, cohort-selection for conducting retrospective research using EHR amongst others. These clinical entities can be categorized into several classes. They can also have attributes that can greatly influence the implied semantics. For example, in the phrase "No orthopnea" in the patient's clinical notes, orthopnea will be flagged as a named entity – i.e. a problem. However, in the absence of a mechanism to indicate the negation in the phrase, any higher-level consumers of the named entities would have an incomplete and sometimes incorrect notion of the named entity. Figure 1 illustrates the attributes of a typical entity span (left) with an example from the annotated dataset - the i2b2 2012 challenge [1] (right), wherein it is indicated that a patient was tested for alkaline phosphate.

Identifying the type and span of named entities are the two predominant aspects of NER. This is also true for negation detection. While identifying the span first and the attributes such as the type and polarity second (whether the entity has been negated etc.), would be ideal, it is intuitive that the type of the named entity is greatly intertwined with the span and it is for this span that the other attributes need to be assigned.

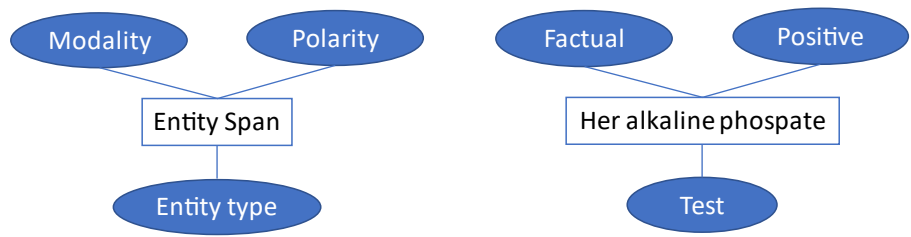

Figure 1: Attributes of an entity

Models that use Bidirectional LSTMs to encode input sequences have dominated the NER space in recent research[2][3][4]. While models that used Bi-LSTMs in conjunction with CNNs[5] did well when they were first published (F1 of 90.69 without external labelled data on CoNLL-2003 test set), the BiLSTM CRF architecture [2] (F1 of 90.94 without external labelled data on CoNLL-2003 test set) and its variants have gained popularity for Named Entity Recognition and subsequently clinical concept extraction [3][4] over the past few years. More recently, models using BiLSTMs as encoders and a combination of feed-forward neural networks and biaffine classifiers have found success in identifying named entities along with nested named entities (Eg. [[Prostate] cancer])[6]. OusiaNER [7] used BERT with a BiLSTM CRF architecture for span identification in parallel with logistic regression for fine-grain multi-label named entity recognition. It then trained this model on large silver-standard corpora (Wikipedia and OntoNotes) and finally finetuned it on a smaller dataset to achieve type labelling.

The concept of self-attention [8] proved to be a game-changer in the field of NER and the BERT[9] model that used self-attention and was finetuned on the NER tagging task achieved an F1 score of 92.8 on the CoNLL-2003 dataset. Alsentzer et. al. [10] used a similar approach and trained BERT vectors on the clinical notes available in the MIMIC-III dataset [11] and additionally finetuned them on the clinical concept recognition task based on the annotated data available in the i2b2 2010/VA [12] and i2b2 2012 [1] challenges' datasets to achieve F1 scores of 87.8 on the i2b2 2010, while Si et.al. used BERT vectors trained on the MIMIC-III dataset within a BiLSTM (Bi-directional Long Short Term Memory) CRF (Conditional Random Field) model to achieve an F1 score of 90.25.

Initially, rule-based algorithms achieved great success in detecting negation in text. NegEx [13] was one of the early examples where clinical terms in text that could be found in the UMLS ontology[14] were replaced by their ICD10 code (International Statistical Classification of Disease and Related Health Problems, $10^{th}$ revision code) to assist with negation target scope detection and regular expression searches were used to identify negation cue words. This was extended to other assertion types such as 'hypothetical', 'historical', or 'experienced by someone other than the patient', by the ConText [15] algorithm. The DEEPEN system [16] incorporated dependency relations into NegEx in order to reduce the number of false positives. The MITRE [17] algorithm was specially formulated for assertion detection in the i2b2 2010/VA [12] challenge. It used a combination of Conditional Random Fields (CRFs), maximum entropy and cue detection to achieve an impressive F1 score of 93 on the i2b2 2010/VA [12] dataset. More recently, negation detection algorithms based on transformers have emerged. Khandelwal et. al. [18] used transformer-based approaches to obtain SOTA results (at the time of publication) on negation scope detection on the BioScope and the SFU Review corpora.

Bhatia et. al. [19] were one of the first to jointly solve the NER task with the negation detection task by treating it as a multitask problem by leveraging the power of teacher forcing in conjunction with an LSTM in the decoder and a Bi-directional LSTM encoder.

Similar to [19], we address the tasks of tagging sequences of text with multiple sets of labels (specifically, for named entities, polarities and optionally modalities) jointly in every pass. However, while [19] appears to focus on tagging only negative findings for polarity, we have focused on tagging all types of polarity and modality where annotated data is available. While determining attributes such as whether a concept is 'hypothetical' or 'intended for someone other than the relevant subject' are important, negative findings are both more frequent [19] and bear a profound impact on the intended meaning of the clinical concepts. Therefore, in this research, we have evaluated the performance of the algorithm on both, the identification of negative clinical concepts exclusively as well as the labelling of the attributes of the identified concepts as one of several possible labels jointly with the algorithm's ability to identify the type of named entities. One of the key challenges with a multi-label tagging problem is that some sequence sets may have a higher degree of class imbalance than other sets. For example, in the annotated i2b2 2010/VA dataset, while the 'other' tag or 'O', which applies to all words that are not part of a named entity may be predominant, the number of tags that belong to the 'Problem'/'Treatment'/'Test' or others are comparable. However, when it comes to tagging the polarity of the tags, only the named entities tagged as problems are assigned a polarity and within those annotated entities, fewer than half have a negative connotation. This class imbalance needs to be accounted for, both while designing a training model and during evaluation. In light of this problem, macro-averaging F1 scores would be a suitable metric for evaluating polarity tagging since micro-averaging will favour the overrepresented class. (Additional information on micro-averaging in the context of multi-label classification is available in the Appendix.). Our best model for NER, the BiLSTM n-CRF model achieved a span based F1 measure of 0.894 for NER and an macro-averaged F1 score of 0.632 for polarity detection on the i2b2 2010 dataset. In the case where only negative clinical concepts are tagged for polarity tagging, the BiLSTM CRF Smax-TF model attained a macro-averaged F1 score of 0.924 for polarity detection, with a span-based F1 score of 0.874 for NER on the same dataset. This architecture also produced models that yielded superlative performance in modality tagging on the i2b2 2012 dataset with a macro-averaged F1 of 0.501, about 0.274 points higher than the BiLSTM n-CRF model.

**BACKGROUND**

In this section, we discuss the related concepts that provide a background to the dual entity and attribute tagging architectures proposed in this paper: namely, tagging conventions and Named Entity Recognition algorithms. Additionally, the benchmarking datasets and sample data are also described here.

In our research, we follow the BILOU tagging scheme which has been used in recent research papers such as [20] [21], where: B - denotes the '**B**eginning', I - denotes the words '**I**nside', L - denotes the '**L**ast' word of a multi-word named entity span, O - denotes words that fall '**O**utside' any named entity span and U - denotes '**U**nit' named entities or named entities comprised of a single word.

For example, a single word named entity of type 'problem' is denoted by 'U-problem'.

The BILOU scheme has been found to be superior to the BIO tagging scheme [22] and is sufficient for the purpose of this research as we are not exploring tagging nested sequences.

Our first architecture uses the BiLSTM CRF model, which is illustrated in Figure 2. The word embedding sequences of words in a sentence are passed to a bi-directional Long Short Term Memory network (LSTM) which gives the left and right context of the word, $l_i$ and $r_i$ respectively. These are then concatenated to form $c_i$ and passed as input to the Conditional Random Field (CRF) layer. The CRF layer is responsible for modelling the rules that result in a valid sequence of tags (Eg. I-Test cannot follow B-Problem) that would be impossible to model using independence assumptions. It does this by assigning transition scores based on the training data [2] as follows: if $X = (x_1, x_2, ..., x_n)$ is an input sequence of words and there exist $k$ distinct sequence tags, an $n \times k$ matrix of scores from the bidirectional LSTM, $\boldsymbol{P}$ is defined, where $P_{i,j}$ corresponds to the score of the $j^{th}$ tag for the $i^{th}$ word in the sentence. For every sequence of tag predictions, $y = (y_1, y_2, ..., y_n)$, the following score term is defined:

$$s(X, y) = \sum_{i=0}^{n} A_{y_i, y_{i+1}} + \sum_{i=1}^{n} P_{i, y_i}$$

where $\boldsymbol{A}$ is a matrix of transition scores such that $A_{i,j}$ represents the score of a transition from the $i^{th}$ tag to the $j^{th}$ tag. Start and end tags, $y_0$ and $y_n$ respectively are added to the set of possible tags making $A$ a square matrix of size $k+2$ [2]. In practical terms, this transition matrix allows the model to inherently discourage illegal transitions – For example, a transition from a 'B-Problem' tag to an 'O' or an 'L-Test'.

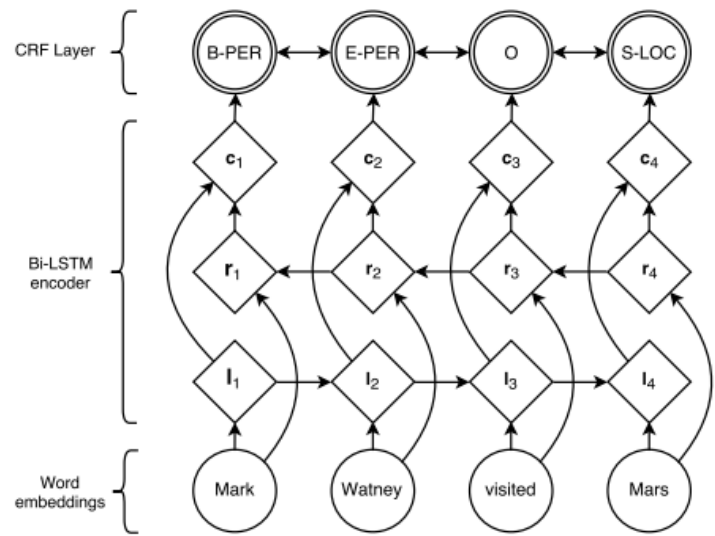

Figure 2: A pictorial representation of the BiLSTM – CRF model [2]

The second model that is relevant and also our baseline is the Conditional Softmax Decoder proposed by Bhatia et. al. [19]. They have proposed 3 architectures to jointly address the task of NER and negation detection. Out of these, the model with the best F1 score in both NER and negation detection is the Conditional Softmax Decoder model and is described in Figure 3. This model also used a Bidirectional LSTM as an encoder – The output of this encoder (Denoted by $h_{enc}$ in Figure 3) is concatenated with the label encoding of the extracted entity label and passed on to a multi-task unidirectional LSTM. During training, the encoding of the gold labels are used (Teacher Forcing [23]) and while testing, the label predicted in the previous time step is used as the extracted entity label. This is then passed through a multi-task LSTM network and the output from the LSTM is first passed through a Softmax decoder that tags named entities. The Softmax output from entity extraction is then passed on to another Softmax layer along with the output from the decoder LSTM in order to predict negation. Passing the output of the entity extraction layer to the negation Softmax allows the layer to learn the distribution of attributes such as negation in relation to the named entities.

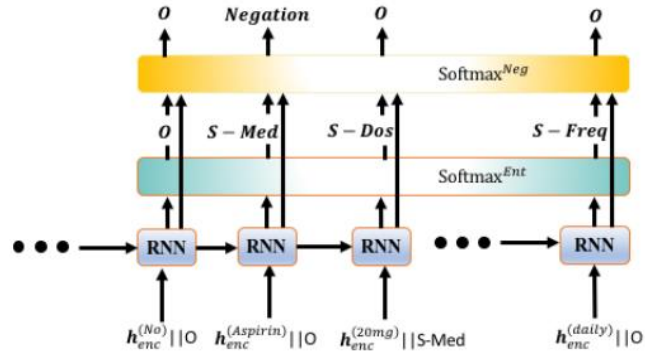

*Figure 3: Conditional Softmax decoder model* [19]

The equations summarizing this architecture are:

$$\hat{y}_t^{Ent}, \text{SoftOut}_t^{Ent} = \text{Softmax}^{Ent}(W^{Ent} o_t + b^s)$$

$$\hat{y}_t^{Neg} = \text{Softmax}^{Neg}(W^{Neg}[o_t, \text{SoftOut}_t^{Ent}] + b^s)$$

Where $o_t$ is the encoded input text, $\hat{y}_t^{Ent}$ and $\hat{y}_t^{Neg}$ represent the entity type and whether it is negated respectively, $\text{Softmax}^{Ent}$ and $\text{Softmax}^{Neg}$ represent the respective Softmax decoders and $W^{Ent}$ and $W^{Neg}$ represent the weights of the respective linear layers and $b^s$ is the bias term. $\text{SoftOut}_t^{Ent}$ is the Softmax output for the entity at time step t. While the Conditional Softmax only addressed negation in clinical text, we have extended this to other attributes such as 'hypothetical', 'associated with someone else' etc.

**METHODOLOGY**

The task of identifying named entities and associated attributes is a sequence tagging problem or a classification task. We formulate this problem of joint entity recognition and attribute detection as a multi-sequence tagging problem. We use BILOU tags to encode the gold sequences passed to the trainer when we are investigating span identification since they encapsulate span information. Since the type of the named entity is tightly coupled with the span, we always prefix the gold tag sequence for NER with BILOU prefixes. The BILOU tagging format is used to tag the input sequence for the modality tasks as well. However, for polarity tagging, we have used the BILOU tagging format for architectures with CRF based decoders (BiLSTM n-CRF and BiLSTM n-CRF-TF architectures) and the un-prefixed format for the Softmax based decoder architectures (Conditional Softmax and BiLSTM CRF-Smax-TF architectures). To accommodate for this, a new exact-matching span based F1 measure evaluation method was developed for the un-prefixed tags generated by the BiLSTM n-CRF and BiLSTM n-CRF-TF architectures – This method simply looks for unbroken sequences of the un-prefixed polarity tags in question and ensures that it matches exactly with the span of the corresponding named entity. The macro-averaged F1 measure has been endorsed in recent literature [24][25] for polarity detection and we have used the same for negation detection. These details are summarized in Figure 4, wherein, if more than one method of evaluation exists and one of them is better suited to the tagging scheme, the preferred method is highlighted in bold. Our architectures do not use data specific feature engineering or supplementary resources such as ontologies or dictionaries and are therefore more generalizable. Further, the word vectors used in this research – the Bio+Clinical BERT [10] vectors have been made publicly available - this makes these models easily reproducible. It is worth noting that the architectures described below use the Bio+Clinical BERT [10] vectors as is and do not finetune them.

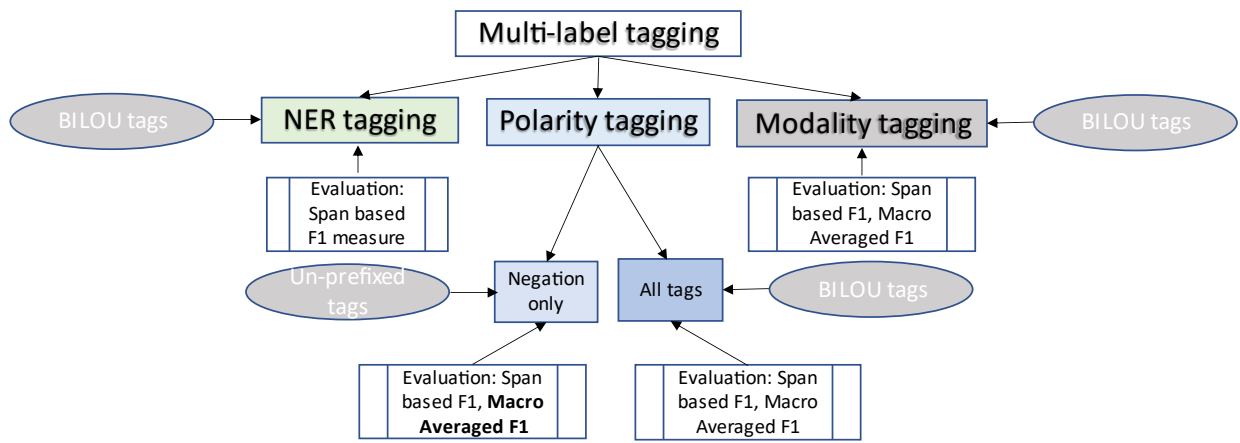

*Figure 4: A summary of the processes and encoding used for the various tagging sequences.*

**Dataset**

For all the proposed architectures and the baselines, we use the datasets from the i2b2 2010/VA [12] and i2b2 2012 [1] challenges which have become the go-to datasets for research in clinical concept recognition [3][4] [10] [19][21][26] and many other tasks, since they are freely available and annotated with named entities, polarities and other information. They contain data where the named entities, polarities and other attributes have been annotated by multiple human annotators. Only a part of the i2b2 2010/VA challenge dataset has been made available since the challenge and we use the existing available split of 170 records for training and 256 records for testing in the i2bb2 2010/VA dataset that is maintained in most current research [19][3][27]. The i2b2 2010/VA challenge provides assertion annotations only for concepts of type "problem". The i2b2 2012 temporal relations challenge data includes 310 discharge summaries (190 records available for training and 120 for testing) consisting of 178000 tokens. These details are summarized in Table 1.

| Dataset | Annotations | No. of disch. sum. | Comments |
|---|---|---|---|
| i2b2 2010/VA [12] | Named Entities, Assertion (Polarity) | 426 (170 training/256 testing) | **Provides assertion annotations only for concepts of type "problem".** Named Entity types: Problem, Treatment, Test. Assertion type: Present, Absent, Possible in the patient, Conditionally present in the patient under certain circumstances, Hypothetically present in the patient at some future point, Mentioned in the patient report but associated with someone other than the patient. |
| i2b2 2012 [1] | Named Entities, Polarity, Modality | 310 (190 training/120 testing) | **Provides polarity and modality information for all identified entity spans.** Named Entity types (taken from [1]): <br>• clinical concepts (problems, tests, and treatments) <br>• clinical departments (such as 'surgery' or 'the main floor') <br>• evidentials (i.e., events that indicate the source of the information, such as the word 'complained' in 'the patient complained about …') <br>• occurrences (i.e., events that happen to the patient, such as 'admission', 'transfer', and 'follow-up'). <br>Polarity types: Positive or Negative <br>Modality Types: Whether an event actually happens, is merely proposed, mentioned as conditional, or described as possible |

*Table 1: Dataset information*

**Baselines**

In order to reproduce the Conditional Softmax Encoder, we followed the configuration and architecture described by Bhatia et.al. [19] closely and implemented the model. As recommended in [19], Glove[28] is used to provide the word embeddings that are consumed by the encoder. For the NegEx baseline, we use negspacy to examine the polarity assigned by the NegEx algorithm on the gold NER spans.

**BiLSTM-nCRF**

In this paper we propose 3 architectures to perform entity and attribute detection – all of these architectures output multiple tag-sequences for an input string. Owing to the success of BiLSTM-CRF models for clinical concept extraction in recent times [3][4], we begin with a model that uses a BiLSTM-CRF tagger to produce each of the sequences. The model uses a shared BiLSTM encoder. We employ a CNN for character level encoding which produces a character level embedding of dimension 16. This is then passed to a BiLSTM encoder with 512 hidden units along with the word embedding from the Bio+ClinicalBERT [10] vectors.

For the BiLSTM encoder, we use the same LSTM configuration as [2], [29] who have in turn based it on [30]:

$$i_t = \sigma(W_{xi}x_t + W_{hi}h_{t-1} + W_{ci}c_{t-1} + b_i)$$
$$f_t = \sigma(W_{xf}x_t + W_{hf}h_{t-1} + W_{cf}c_{t-1} + b_f)$$

$$c_t = (1 - i_t) \odot c_{t-1} + i_t \odot \tanh(W_{xc}x_t + W_{hc}h_{t-1} + b_c)$$
$$o_t = \sigma(W_{xo}x_t + W_{ho}h_{t-1} + W_{co}c_t + b_o)$$
$$h_t = o_t \odot \tanh(c_t)$$

where $\sigma$ is the element-wise sigmoid function, $\odot$ the element-wise product, $c$ the cell state, $i, f$ and $o$ are the input, forget and output gates respectively and are of the same size as the hidden state vector $h$. $X$ is the input sequence denoted by $X = (x_1, x_2, ..., x_n)$ and $t$ is the timestep or in this case, the index of the word being currently processed. The meaning of the weight matrix subscripts are intuitive - for example, $W_{hi}$ is the hidden-input gate matrix, $W_{xo}$ is the input-output gate matrix etc. All cell to gate weight matrices are diagonal, so element m in each gate vector only receives input from element m of the cell vector [29].

The output of this encoder is then passed to the CRF decoders, one for the entity type and one for each attribute to be tagged. The matrix of scores output by the bidirectional LSTM network, denoted by $P$, is of size $n \times k$, where $n$ is the sequence length and $k$ is the number of distinct tags. $P_{i,j}$ represents the score of the $j^{th}$ tag for the $i^{th}$ word in a sentence. For a sequence of predictions $y = (y_1, y_2, ..., y_n)$, if $A$ denotes the matrix of transition scores where $A_{i,j}$ represents the score for the transition from tag $i$ to tag $j$, the score for the tag sequence is modelled as follows, similar to [2]:

$$s(X, y) = \sum_{i=0}^{n} A_{y_i, y_{i+1}} + \sum_{i=1}^{n} P_{i, y_i}$$

A Softmax over all possible tag sequences gives the probability for the sequence $y$.

$$p(y|X) = \frac{e^{s(X,y)}}{\sum_{\tilde{y} \in Y_X} e^{s(X,\tilde{y})}}$$

Here, $Y_X$ denotes all possible tag sequences. The objective then becomes maximizing the log probability of the correct tag sequence while training.

$$log(p(y|X)) = s(X, y) - \log\left(\sum_{\tilde{y} \in Y_X} e^{s(X,\tilde{y})}\right) \qquad 1$$

$$y^* = argmax_{\tilde{y} \in Y_X} s(X, \tilde{y}) \qquad 2$$

Dynamic programming (Viterbi algorithm [31]) is used in Eq. 2 to deduce the final tag sequence for the NER and other attributes. The final loss function is a weighted sum of the individual negative log losses from each CRF decoder. This architecture is illustrated in Figure 5. The portion of the architecture that tags modality is dotted as these tags are only available in the i2b2 2012 dataset and not in the i2b2 2010/VA dataset.

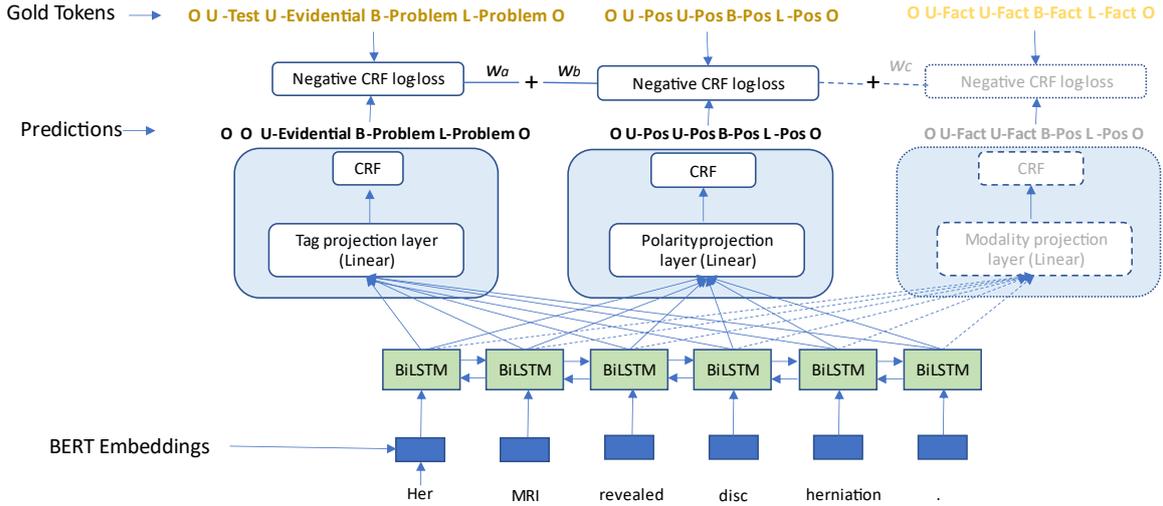

*Figure 5: The BiLSTM n-CRF model*

**BiLSTM CRF-Smax-TF**

Our next model, the BiLSTM CRF-Smax-TF is represented in Figure 6. It uses the same shared BiLSTM encoder (with the same input processing) and Bio+Clinical BERT vectors [10] as the BiLSTM n-CRF model but different decoders.

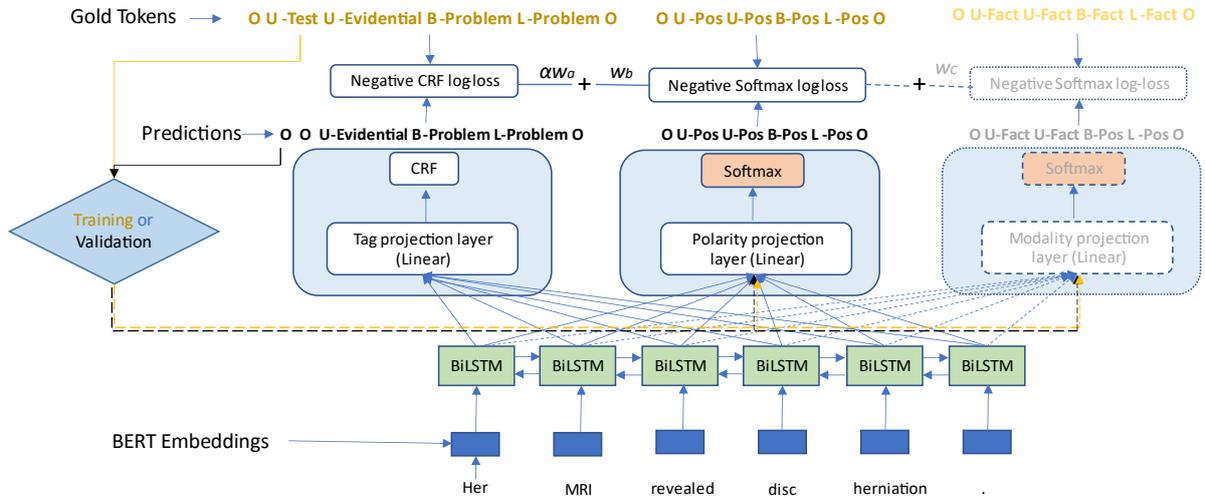

*Figure 6: BiLSTM CRF-Smax-TF*

The NER is carried out by a CRF layer employing the Viterbi algorithm as in the BiLSTM n-CRF model but the attribute tagging is achieved using a simple Softmax layer since it has achieved success in detecting negation [19][32]. Similar to Bhatia et al. [19] the tag sequence from the NER layer is passed to the polarity (or attribute) detection layer so that the architecture is more responsive to selective annotations in the data, such as that in the i2b2 2010 where only the entities of type 'problem' have polarities annotated. This is represented by the dashed yellow and grey lines in Figure 6.

Additionally, while teacher forcing has traditionally been a concept exclusive to RNNs, we adapt it and pass the ground truth labels for the whole batch while training and NER predictions for the batch during validation. The loss is again a weighted combination of the negative log loss of the CRF and Softmax layers, with the CRF loss scaled to match the loss in the Softmax layer.

**BiLSTM n-CRF-TF**

The BiLSTM n-CRF-TF model takes a best of both worlds approach and incorporates teacher forcing into the n-CRF architecture. All tagging sequences are generated by BiLSTM CRF layers using principles of dynamic programming such as the Viterbi algorithm, as in the case of the BiLSTM n-CRF. However, the output sequence from the NER tagger/gold NER tags (depending on whether the model is being validated or trained respectively) is concatenated to the input to the other tagging layers, thereby implementing teacher-forcing. This is shown in Figure 7. The key differences between the BiLSTM CRF-Smax-TF and BiLSTM nCRF-TF architectures are highlighted in orange/grey.

The key difference between the BiLSTM n-CRF-TF model and the BiLSTM CRF-Smax-TF model is the use of CRFs for attribute prediction. It is expected that this will allow for better span prediction of the attributes.

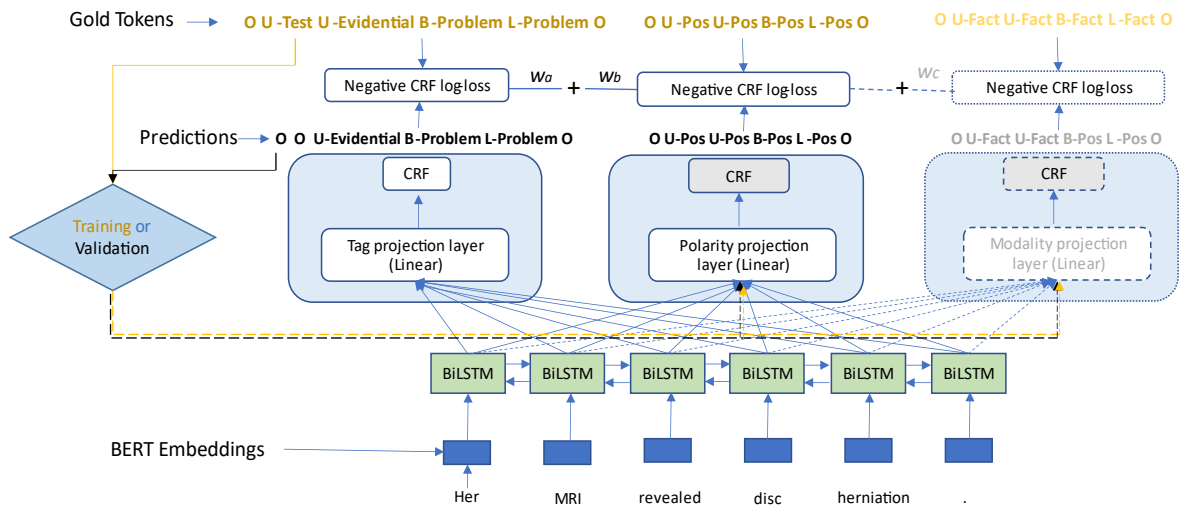

*Figure 7: BiLSTM nCRF-TF*

In addition to the above methods, other prototypes were also developed, but failed to yield promising outcomes. One such prototype was that of a bootstrapping model that fed back the tagged polarity sequence to the NER tagger - this extended the teacher forcing concept to leverage the superiority of the polarity tagger in certain scenarios. These architectures, however, did not yield encouraging results and are therefore not discussed here.

**RESULTS AND DISCUSSION**

In this section, we present the results of our 3 proposed architectures on NER, polarity and optionally modality tag sequence prediction. To set a more generic guideline, the results for NER and polarity detection, summarized in Table 2 serve as unstructured baselines:

| Method | Dataset | NER-F1 | Polarity-F1 | Comments |
|---|---|---|---|---|
| Hidden semi-Markov Model(De Bruijn et. al.[24]) | I2b2 2010/VA | 0.852 | 0.936 | Used semi-Markov Hidden Markov Models for NER and SVMs for assertion(polarity) detection |
| BiLSTM-CRF (R. Chalapathy and Piccardi [3]) | I2b2 2010/VA | .839 | - | Seminal paper that used BiLSTM CRF to perform NER in the clinical context on i2b2 2010. Does not perform polarity detection. |
| Independent NER (Lample et al. [2]) | I2b2 2010/VA | 0.848 [19] | - | Seminal paper that used BiLSTM CRF to perform NER – however not in the clinical context. Does not perform polarity detection. |
| Bio+Clinical BERT(Alsentzer et al. [10]) | I2b2 2010/VA | 87.2 | - | Trained and finetuned a BERT model on part of the text available in the MIMIC III [11] dataset. |
| Si et. al. [4] | I2b2 2010/VA | 0.903 | - | Trained a BERTLARGE model on the clinical notes available in the MIMIC III [11] dataset from scratch. Do not make their embeddings public. No polarity detection. |
| KGQA(Banerjee et. al [33]) | I2b2 2010/VA | 0.9267 | - | SOTA Clinical concept recognition. This model also does not detect negation or other attributes of the named entities. |
| Xu et.al. [34] | i2b2 2012 | 0.791 | 0.791 | Obtained the best F1 scores for both NER and polarity detection in the i2b2 2012 challenge itself, using a dosage-unit dictionary, CRFs, SVMs, a rule-based negation detection system and intensive feature engineering. |
| Bio+Clinical BERT(Alsentzer et al. [10]) | i2b2 2012 | 78.9 | - | Trained and finetuned a BERT model on part of the text available in the MIMIC III [11] dataset. |
| Si et. al. [4] | i2b2 2012 | 0.809 | - | Trained a BERT$_{LARGE}$ model on the clinical notes available in the MIMIC III [11] dataset from scratch. Do not make their embeddings public. . No polarity detection. |
| KGQA(Banerjee et. al [33]) | i2b2 2012 | 0.8398 | - | SOTA Clinical concept recognition. This model also does not detect negation or other attributes of the named entities. |

*Table 2: Generic Baselines*

We use the publicly released Bio+Clinical BERT [10] which has been trained on only a portion of textual data from the MIMIC III [11] dataset that was used to train the BERT models in [4]. This is because Si et. al. do not make their model public. The SOTA for the i2b2 2010/VA challenge for exclusive NER using Bio+Clinical BERT presently stands at an F1 score of 87.2 [10] while that for i2b2 2012 is 78.9. Since Bhatia et. al. [19] have conducted one of the very few research projects that handle the joint task of entity and attribute (negation) detection, we benchmark our task against their best performing model, the Conditional Softmax Decoder. All our experiments and baselines have been implemented using Allennlp [35]

The results of our research are presented in tables 3 through 9. The variation in the polarity tagging schemes, between the BILOU and the un-prefixed tags, implies that the macro-averaged F1 scores for the BiLSTM n-CRF and BiLSTM n-CRF-TF architectures are more fine-grained as the categories are prefixed and therefore have fewer training samples.

For the polarity detection task, we conduct experiments by passing both the BILOU prefixed tags as well as the absolute tags as suggested in the 'Other info' column. Negation only indicates that unprefixed tags of only the negative/absent type (such as 'NEG') are passed to the model during polarity training – i.e. tags indicating other semantics such as positive/present, hypothetical etc. do not form part of the training data and are marked as the 'O' tag. 'BILOU-Neg only' indicates that BILOU-prefixed polarity tags of type negative/absent only are passed. The Span based F1 scores are calculated based on an exact span match. However, in keeping with common practice for exact span matching, we compute micro-averaged results which may favour the overrepresented class. The macro-averaged results are also shown for polarity and modality studies.

We highlight the models trained on the negation polarity type only in blue and discuss the best performing architectures in both categories (Negation only/ all polarity types). The first row in both categories in every table represents the results obtained for our implementation of the Conditional Softmax baseline. The following 3 rows represent our 3 architectures. Table 3 and Table 4 compare the performance of our 3 models, namely the BiLSTM n-CRF, BiLSTM CRF-Smax-TF and BiLSTM n-CRF-TF models with that of the Conditional Softmax Decoder for the NER and Polarity detection tasks respectively on the i2b2 2010/VA dataset.

| Method | Other parameters | Accuracy - NER | Precision | Recall | Span-based F1 |
|---|---|---|---|---|---|
| Conditional Softmax [19] | BILOU | 0.947 | 0.858 | 0.850 | 0.854 |
| BiLSTM n-CRF | BILOU | 0.962 | 0.887 | 0.903 | **0.894** |
| BiLSTM CRF-Smax-TF | BILOU | 0.953 | 0.862 | 0.881 | 0.871 |
| BiLSTM n-CRF-TF | BILOU | 0.953 | 0.865 | 0.879 | 0.872 |
| Conditional Softmax [19] | Negation only | 0.946 | 0.862 | 0.842 | 0.852 |
| BiLSTM n-CRF | BILOU+Neg only | 0.962 | 0.889 | 0.899 | **0.896** |
| BiLSTM CRF-Smax-TF | Negation only | 0.954 | 0.873 | 0.875 | 0.874 |
| BiLSTM n-CRF-TF | BILOU+Neg only | 0.953 | 0.874 | 0.867 | 0.870 |

*Table 3: Performance of various classifiers on the NER task on the i2b2 2010 dataset*

| Method | Other parameters | Accuracy | Precision | Recall | Span-based F1 | Macro-avg P | Macro-avg R | Macro-avg F1 |
|---|---|---|---|---|---|---|---|---|
| Conditional Softmax [19] | BILOU | 0.967 | 0.747 | 0.768 | 0.757 | 0.553 | 0.527 | 0.532 |
| BiLSTM n-CRF | BILOU | 0.974 | 0.823 | 0.841 | **0.832** | 0.770 | 0.630 | 0.632 |
| BiLSTM CRF-Smax-TF | BILOU | 0.972 | 0.772 | 0.800 | 0.786 | 0.718 | 0.626 | **0.638** |
| BiLSTM n-CRF-TF | BILOU | 0.967 | 0.773 | 0.810 | 0.791 | 0.649 | 0.502 | 0.525 |
| Negex [13] | Negation only | - | 0.9 | 0.78 | 0.84 | 0.9 | 0.78 | 0.84 |
| Conditional Softmax [19] | Negation only | 0.995 | 0.883 | 0.804 | 0.842 | 0.946 | 0.895 | 0.919 |
| BiLSTM n-CRF | BILOU-+Neg only | 0.996 | 0.88 | 0.895 | **0.888** | 0.890 | 0.888 | 0.889 |
| BiLSTM CRF-Smax-TF | Negation only | 0.995 | 0.784 | 0.899 | 0.838 | 0.899 | 0.952 | **0.924** |
| BiLSTM n-CRF-TF | BILOU-+Neg only | 0.995 | 0.840 | 0.879 | 0.859 | 0.878 | 0.854 | 0.864 |

*Table 4: Performance of various classifiers on polarity detection on the i2b2 2010 dataset*

The composition of the loss function does not impact the performance of the model in the n-CRF based architectures (**Error! Reference source not found.**). But for the BiLSTM CRF-Smax-TF, NER log loss is scaled down to be comparable to the polarity loss in order to effectively backpropagate losses through the NER and attribute tagging networks. These values have been chosen empirically, with the NER given precedence. The i2b2 2010/VA dataset only annotates polarity for entities of type 'Problem' – This causes a strong class imbalance. Additionally, when we focus on identifying only tags that are negated, during training, we classify all polarity/assertion tags other than 'absent' as non-attributes which causes only a minority of named entities to have a polarity annotated.

From Table 3 it is seen that the BiLSTM n-CRF model is the best performing model for NER with BiLSTM CRF-Smax-TF being a close second, not more than 2 percentage points away in all categories. For the polarity detection task (Table 4), the BiLSTM n-CRF again performs best in the span based F1 measure category. However, for negation detection only, the BiLSTM CRF-Smax-TF performs the best. The introduction of teacher forcing to the BiLSTM n-CRF model proves disadvantageous for both NER and polarity tagging on the i2b2 2010/VA dataset. The CRF layer is initialised with the transition score matrix using the BILOU tags of the corresponding attribute (NER/polarity/modality). Traditionally, this layer is then trained to compute the path with the least score using dynamic programming, by factoring in the 'emission' scores emerging from the BiLSTM encoder (Left/Right context) with the transition scores. However, when tag embeddings which are differently annotated to the ground truth (i.e., i2b2 2010/VA dataset annotates problem, treatment and test named entities but provides polarity information for the problem type alone) are concatenated to the output of the BiLSTM encoder, it causes the performance of the polarity tagging CRF layer and therefore the whole model to be slightly compromised. The Softmax model is more resilient to data with annotation imbalance than the CRF model with dynamic programming.

Table 5 and Table 6 compare the performance of the different models on the i2b2 2012 dataset while considering all the polarity tags as well as Negative tags only. For the BiLSTM n-CRF based experiments, the proportion of NER to polarity losses used in the composite loss function do not have a major influence on the resulting NER and polarity tagging F1 scores. However, for the BiLSTM CRF-Smax-TF, these coefficients were 0.0002 and 1 respectively. A detailed analysis of the loss function composition and its impact on NER and polarity tagging can be found in **Error! Reference source not found.**

The BiLSTM n-CRF-TF shows the best NER classification for the first category while the BiLSTM CRF-Smax-TF has the best F1 measure for the Negation-only category. However, the 3 architectures perform very similarly in the NER evaluation, as expected. The polarity tagging results are different to i2b2 2010/VA, however, with BiLSTM n-CRF -TF having the highest F1 scores in both categories for span based F1 metrics but Softmax based architectures give the best macro-averaged F1 scores and are more accurate if spans are ignored.

| Method | Other parameters | Accuracy | Precision | Recall | Span-based F1 |
|---|---|---|---|---|---|
| Conditional Softmax [19] | BILOU | 0.870 | 0.745 | 0.755 | 0.750 |
| BiLSTM n-CRF | BILOU | 0.897 | 0.807 | 0.806 | 0.806 |
| BiLSTM CRF-Smax-TF | BILOU | 0.896 | 0.794 | 0.816 | 0.805 |
| BiLSTM n-CRF-TF | BILOU | 0.898 | 0.808 | 0.809 | **0.808** |
| Conditional Softmax [19] | BILOU | 0.871 | 0.752 | 0.754 | 0.753 |
| BiLSTM n-CRF | BILOU | 0.896 | 0.807 | 0.808 | 0.808 |
| BiLSTM CRF-Smax-TF | BILOU | 0.899 | 0.814 | 0.808 | **0.811** |
| BiLSTM n-CRF-TF | BILOU | 0.898 | 0.813 | 0.806 | 0.810 |

*Table 5 : Performance of various classifiers on the NER task on the i2b2 2012 dataset*

| Method | Other parameters | Accuracy | Precision | Recall | Span-based F1 | Macro-avg P | Macro-avg R | Macro-avg F1 |
|---|---|---|---|---|---|---|---|---|
| Conditional Softmax [19] | BILOU | 0.910 | 0.773 | 0.793 | 0.783 | 0.874 | 0.854 | 0.864 |
| BiLSTM n-CRF | BILOU | 0.910 | 0.828 | 0.838 | 0.833 | 0.828 | 0.808 | 0.818 |
| BiLSTM CRF-Smax-TF | BILOU | 0.925 | 0.809 | 0.835 | 0.822 | 0.886 | 0.873 | **0.879** |
| BiLSTM n-CRF-TF | BILOU | 0.911 | 0.836 | 0.837 | **0.836** | 0.811 | 0.823 | 0.816 |
| Negex [13] | Negation only | - | 0.6 | 0.84 | 0.7 | 0.6 | 0.84 | 0.7 |
| Conditional Softmax [19] | Neg only | 0.990 | 0.733 | 0.728 | 0.730 | 0.904 | 0.873 | **0.888** |
| BiLSTM n-CRF | BILOU-Neg only | 0.989 | 0.763 | 0.749 | 0.756 | 0.795 | 0.795 | 0.795 |
| BiLSTM CRF-Smax-TF | Neg only | 0.989 | 0.710 | 0.754 | 0.732 | 0.884 | 0.889 | 0.887 |
| BiLSTM n-CRF-TF | BILOU-Neg only | 0.989 | 0.756 | 0.767 | **0.761** | 0.782 | 0.815 | 0.798 |

*Table 6: Performance of various taggers on polarity detection on the i2b2 2012 dataset*

In addition to the above, we extend the BiLSTM n-CRF and the BiLSTM CRF-Smax-TF algorithms to include predicting the modality of named entities. The results are summarised in Table 7, Table 8 and Table 9. For the BiLSTM n-CRF based experiments, the proportion of NER to polarity to modality losses used in the composite loss function was 0.6, 0.2 and 0.2 respectively. Similarly, for the BiLSTM CRF-Smax-TF, these proportions were 0.0002, 1 and 3 respectively. The proportions of the individual losses factored in the composite loss function were empirically evaluated.

| Method | Other parameters | Accuracy | Precision | Recall | Span-based F1 |
|---|---|---|---|---|---|
| Conditional Softmax[19] | BILOU | 0.870 | 0.745 | 0.755 | 0.750 |
| BiLSTM n-CRF | BILOU | 0.893 | 0.803 | 0.815 | 0.809 |
| BiLSTM CRF-Smax-TF | BILOU | 0.896 | 0.803 | 0.796 | 0.799 |
| BiLSTM n-CRF-TF | BILOU | 0.900 | 0.810 | 0.810 | **0.810** |
| Conditional Softmax[19] | BILOU | 0.871 | 0.752 | 0.754 | 0.753 |
| BiLSTM n-CRF | BILOU | 0.895 | 0.804 | 0.808 | 0.806 |
| BiLSTM CRF-Smax-TF | BILOU | 0.899 | 0.812 | 0.792 | 0.802 |
| BiLSTM n-CRF-TF | BILOU | 0.897 | 0.797 | 0.819 | **0.808** |

*Table 7: Performance of various taggers on NER on the i2b2 2012 dataset when modality is included.*

| Method | Other parameters | Accuracy | Precision | Recall | Span-based F1 | Macro-avg P | Macro-avg R | Macro-avg F1 |
|---|---|---|---|---|---|---|---|---|
| Conditional Softmax[19][19] | BILOU | 0.910 | 0.773 | 0.793 | 0.783 | 0.874 | 0.854 | 0.864 |
| BiLSTM n-CRF | BILOU | 0.912 | 0.825 | 0.844 | 0.834 | 0.832 | 0.808 | 0.819 |
| BiLSTM CRF-Smax-TF | BILOU | 0.923 | 0.822 | 0.819 | 0.820 | 0.889 | 0.859 | **0.873** |
| BiLSTM n-CRF-TF | BILOU | 0.912 | 0.836 | 0.836 | **0.836** | 0.817 | 0.816 | 0.816 |
| Conditional Softmax[19] | Negation only | 0.990 | 0.733 | 0.728 | 0.730 | 0.904 | 0.873 | 0.888 |
| BiLSTM n-CRF | BILOU-Neg only | 0.989 | 0.793 | 0.649 | 0.714 | 0.862 | 0.688 | 0.759 |
| BiLSTM CRF-Smax-TF | Negation only | 0.990 | 0.748 | 0.735 | 0.741 | 0.904 | 0.876 | **0.889** |
| BiLSTM n-CRF-TF | BILOU-Neg only | 0.910 | 0.824 | 0.847 | **0.835** | 0.827 | 0.807 | 0.815 |

*Table 8: Performance of various taggers on polarity detection on the i2b2 2012 dataset when modality is included.*

| Method | Other parameters | Accuracy | Precision | Recall | Span-based F1 | Macro-avg P | Macro-avg R | Macro-avg F1 |
|---|---|---|---|---|---|---|---|---|
| BiLSTM n-CRF | BILOU | 0.900 | 0.802 | 0.821 | 0.812 | 0.258 | 0.225 | 0.227 |
| BiLSTM CRF-Smax-TF | BILOU | 0.915 | 0.792 | 0.797 | 0.795 | 0.550 | 0.498 | **0.493** |
| BiLSTM n-CRF-TF | BILOU | 0.903 | 0.815 | 0.815 | **0.815** | 0.512 | 0.254 | 0.277 |
| BiLSTM n-CRF | BILOU-Neg only | 0.899 | 0.801 | 0.820 | 0.810 | 0.300 | 0.218 | 0.218 |
| BiLSTM CRF-Smax-TF | Negation only | 0.915 | 0.800 | 0.792 | 0.790 | 0.568 | 0.496 | **0.501** |
| BiLSTM n-CRF-TF | BILOU-Neg only | 0.902 | 0.807 | 0.829 | **0.818** | 0.497 | 0.299 | 0.318 |

*Table 9: Performance of various taggers on modality detection on the i2b2 2012 dataset when modality is included.*

As in the previous experiments, the BiLSTM CRF-Smax-TF model achieves significantly better scores for macro-averaged polarity while the BiLSTM nCRF -TF model is the best for span-based F1 measuring experiments, affirming the expectation that CRF-based dynamic programming models are excellent at span detection. The modality scores for macro-averaging increase from nearly 0 when modality was not part of the loss function to about 0.5 with BiLSTM CRF-Smax-TF performing the best in for macro-averaged F1 results and BiLSTM n-CRF-TF giving the best scores for span based F1. Additionally, this has not come at a substantial cost to the NER or polarity detection scores.

We have proposed and evaluated 3 architectures, each with its own strengths. From our evaluation, it can be seen that CRF based methods perform well for span detection while the BiLSTM CRF-Smax-TF model performs well when classes are imbalanced. The CRF based methods use a combination of dynamic programming and conditional random fields which intrinsically prevents the formation of illegal tag sequences – This property allows the network to be better at span matching, thereby generating better span based F1 scores. Further, teacher forcing informs the layers responsible for the polarity/modality tagging of the NER spans that were identified/gold labels. We discovered that the Softmax model is more adaptive to recognizing the patterns of selective attribute annotations in the text using the information from teacher forcing(i.e. When annotations are only available for a few entity types). Therefore, teacher forcing is more effective for the BiLSTM CRF-Smax-TF model in the dataset with annotation class imbalance but proves slightly advantageous for the CRF based model when training with data that is balanced.

**RELATED WORK**

While Bhatia et. al. [19] are one of the first and few to solve the NER with the negation detection task, models that jointly accomplish span and sentiment/attribute detection have emerged in other areas of Natural Language Processing in recent times. For example, in the area of sentiment analysis, 'aspect term' extraction and 'aspect sentiment' classification have traditionally been treated as separate tasks. When assessing the review of a product, the 'aspect term' deals with a feature (Eg. processing speed of a PC) discussed within the review and 'aspect sentiment' deals with the sentiment associated with the feature (positive, negative, neutral, conflict etc.). Akhtar et. al [36] propose a method to jointly solve these tasks by using a BiLSTM encoder with self attention to tag the aspect term and a convolution layer above it for aspect classification. This method achieves competitive performance for aspect classification. Wang et. al. [37] propose the use of shared components (encoders, capsule embeddings and attention mechanisms) to exploit the correlation between and jointly address the tasks of aspect detection and sentiment classification. Each aspect category is assigned a capsule that consumes text encoded by a shared RNN and jointly learns to recognize aspect spans and sentiments for that category. They achieved state of the art results at the time of their publication.

Xiong et. al. [38] even apply the principles of machine reading comprehension to accomplish Medical named Entity Recognition (MER) and Medical named Entity Normalisation (MEN) on a spanish tumour morphology challenge dataset called CANTEMIST [39]. The MEN task involvs coding the recognized named entities into a bespoke scheme – most of the coding relies on information contained within the entity itself, some tumour mentions contain a modifier (scientific adjective qualifier for the tumour) not included in the standardized coding scheme (such as ICD) for the specific tumour but still relevant. They achieve significant improvements over pipeline based methods for MEN.

**CONCLUSION**

In this paper, we have proposed 3 architectures for the detection and classification of Named Entities and their Attributes. To the best of our knowledge, our models have resulted in a better overall performance in jointly predicting NER and polarity tags when compared to other recent architectures which attempt to do the same. While the BiLSTM n-CRF model, based on multiple CRFs, one for each span and attribute to be predicted, shows the best results for NER and for span based F1 measures of polarity detection in the i2b2 2010 dataset, the BiLSTM CRF-Smax-TF is a better option to consider when there is a strong class imbalance in the data and we are interested in identifying tags that occur infrequently. The BiLSTM n-CRF-TF offers significantly lower performance for NER in i2b2 2010/VA dataset when compared to the BiLSTM n-CRF model but performs better in the span based evaluations in the i2b2 2012 datasets. However, it is very comparable to the BiLSTM n-CRF model in performance on this dataset, except for span based polarity detection in the negated case making it a candidate for data with class imbalances in one of the tagging schemes. As a future endeavour, research into models that optimally share parts of the decoder networks to predict named entities and attributes together would be beneficial and would result in improved efficiency. Additionally, the application of methods based on sentiment analysis, reading comprehension (as sampled in the related work section) to the problem of joint entity and attribute detection holds great promise as does the use of ontologies such as the Unified Medical Language System [14] to demarcate and identify nested concepts. It is hoped that the cross-fertilisation of ideas from different areas of Natural Language Processing can result in powerful algorithms in the clinical entity and attribute recognition space.

**ACKNOWLEDGEMENTS**

We would like to thank A/Prof. Mark D McDonnell for making the resources needed for our experimentation available.

**APPENDIX A.**

## Micro-averaging in the context of multi-label/multi-class classification

The precision of any classifier is defined as the ratio of the number of samples that were correctly classified (True Positives or TP) to be of a certain class to the total number of samples classified as belonging to the class (True Positives+False Positives or FP): TP/(TP+FP). Recall is the fraction of all samples of a class that get correctly classified as belonging to the class: TP/TP+FN where FN represents False Negatives. F1 scores are defined to be the harmonic mean of precision and accuracy. While micro-averaging, the F1 score for the multi-class classifier is calculated by summing the True Positives, False Negatives and False positives across classes. This process leads to the F1 score for each class being the same. This phenomenon can be explained by the below table similar to that in [40].

| Gold Label | 1 | 2 | 3 | 2 | 3 | 3 | 1 | 2 | 2 |
|---|---|---|---|---|---|---|---|---|---|
| Prediction | 3 | 2 | 1 | 2 | 2 | 3 | 3 | 1 | 2 |

In a system with gold labels and corresponding predictions as in the table above, every time a sample is wrongly classified to belong to a class (False Positive), it is also a False Negative to the class from the gold label (False Negative). If class A is predicted and the true label is B, then there is a FP for A and a FN for B. This causes FP = FN. True Positives have no effect on FP or FN. Therefore,

$$Precision = \frac{TP}{TP + FP}$$

$$Recall = \frac{TP}{TP + FN}$$

Since FP = FN,

$$Recall = \frac{TP}{TP + FP}$$

Therefore,

$$Precision = Recall \tag{1}$$

$$F = \frac{2 * Precision * Recall}{Precision + Recall} \tag{2}$$

Substituting (1) in (2)

$$F = Precision$$

Therefore, for micro-averaging,

$$F = Precision = Recall$$

However, while macro-averaging, the F1 scores are computed for each class individually and averaged.

# APPENDIX B.

## Ablation studies

We present the ablation study done by varying the coefficients of multiplication for the NER and polarity detection in the composite loss function. We limit the study to entity type and polarity tagging only and use span based F1 measure for polarity tagging evaluation for the BiLSTM nCRF architectures but macro averaged polarity for the BiLSTM CRF-Softmax-TF model. We notice that changing the coefficients does not make a profound impact on the performance of the algorithms except in the case of the CRF-Softmax-TF architecture where the NER loss from the CRF is several orders of magnitude larger than the Softmax loss and needs to be scaled down for the effective backpropagation of both losses.

I2b2 2010/VA ablation studies – F1 scores vs loss composition

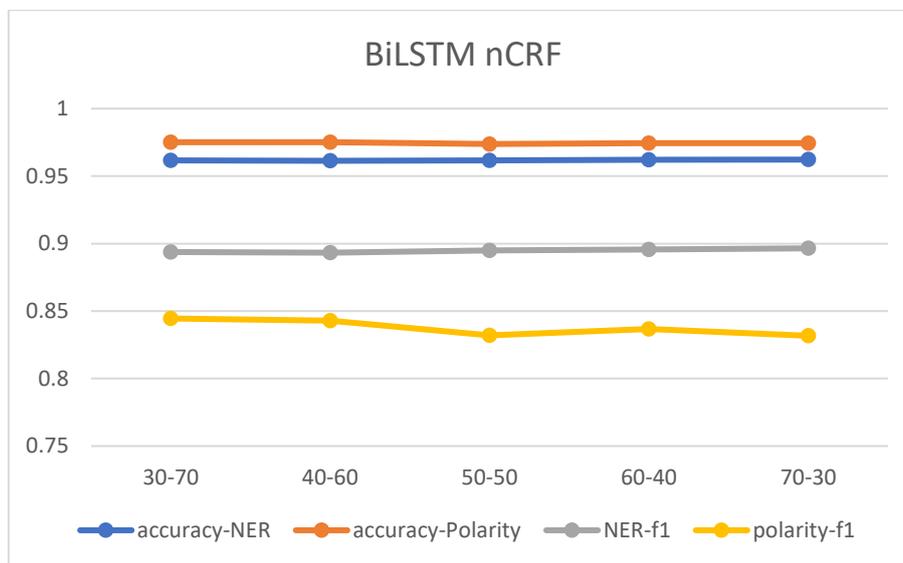

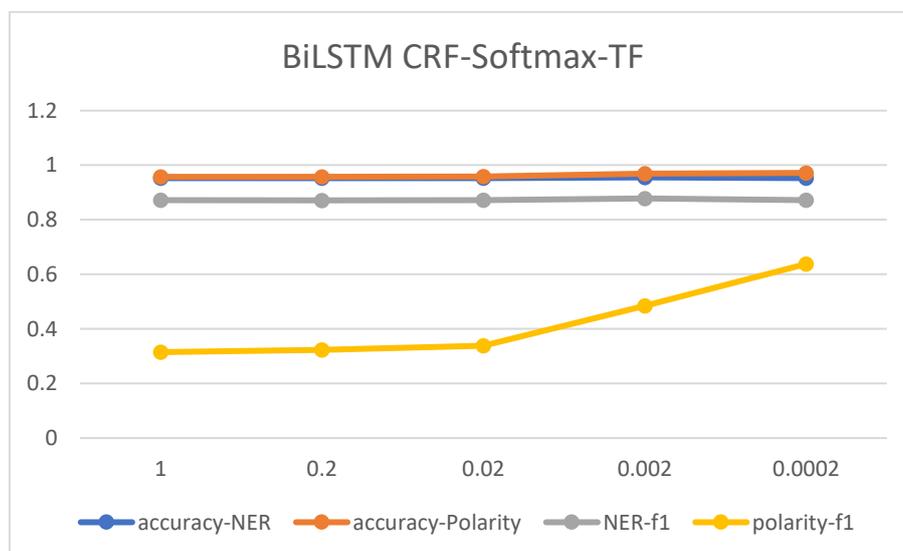

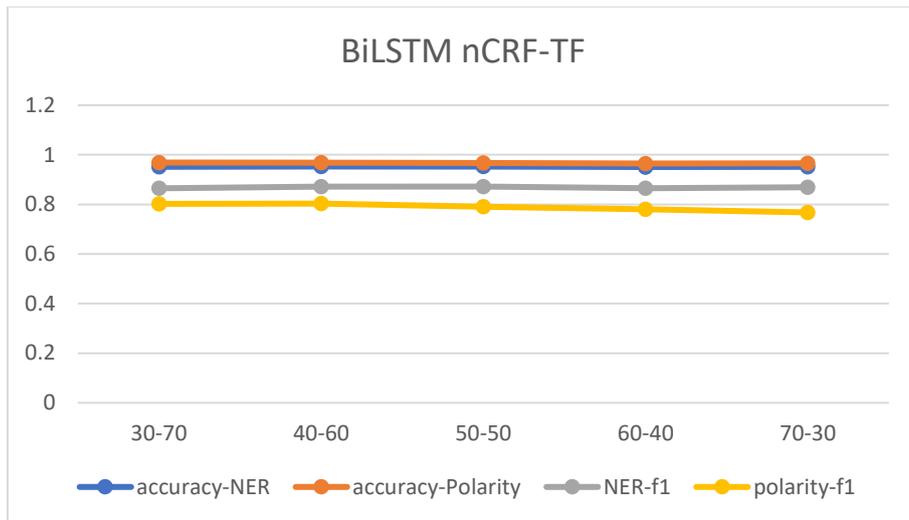

**I2b2 2012 ablation studies – F1 scores vs loss composition**

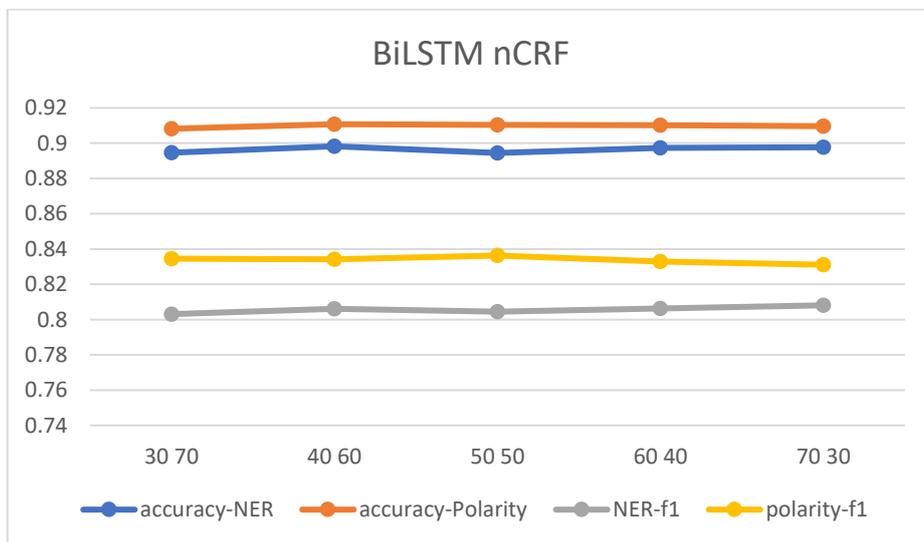

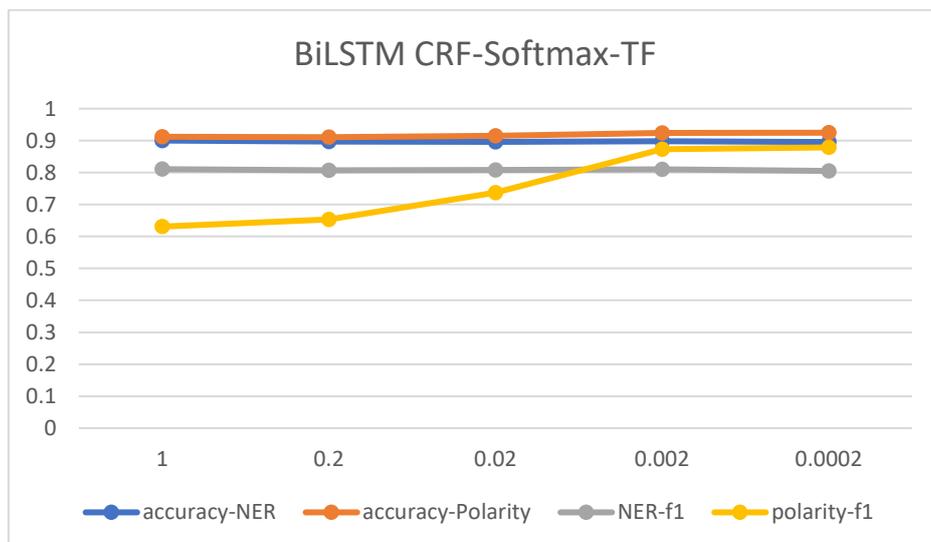

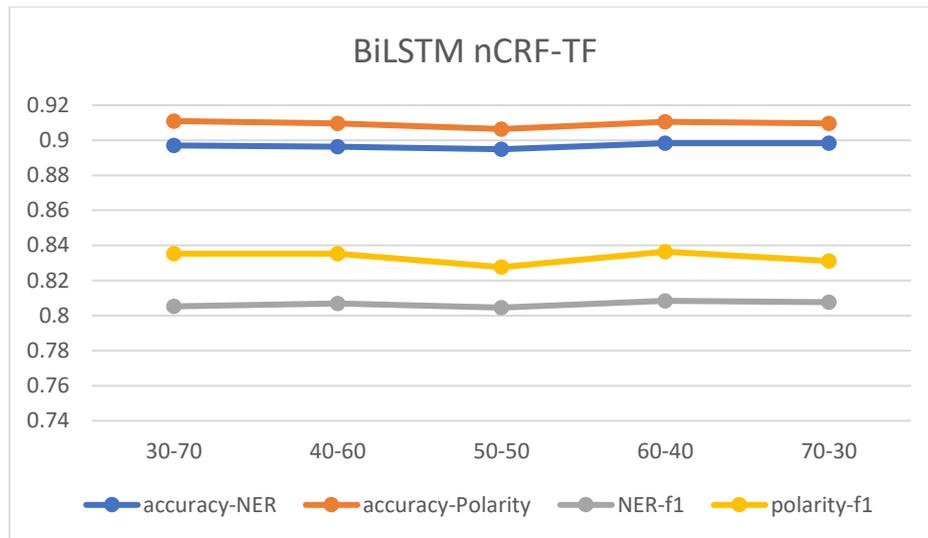

**BIBLIOGRAPHY**

[1]   W. Sun, A. Rumshisky, and O. Uzuner, "Evaluating temporal relations in clinical text: 2012 i2b2 Challenge," *Journal of the American Medical Informatics Association*, vol. 20, no. 5. pp. 806–813, 2013.

[2]   G. Lample, M. Ballesteros, S. Subramanian, K. Kawakami, and C. Dyer, "Neural Architectures for Named Entity Recognition," in *The North American Chapter of the Association for Computational Linguistics: Human Language Technologies (NAACL-HLT)*, 2016.

[3]   R. Chalapathy, E. Z. Borzeshi, and M. Piccardi, "Bidirectional LSTM-CRF for Clinical Concept Extraction," in *Proceedings of the Clinical Natural Language Processing Workshop (ClinicalNLP)*, 2016, pp. 7–12.

[4]   Y. Si, J. Wang, H. Xu, and K. Roberts, "Enhancing clinical concept extraction with contextual embeddings," *J. Am. Med. Informatics Assoc.*, 2019.

[5]   J. P. C. Chiu and E. Nichols, "Named Entity Recognition with Bidirectional LSTM-CNNs," *Trans. Assoc. Comput. Linguist.*, vol. 4, pp. 357–370, 2016.

[6]   J. Yu, B. Bohnet, and M. Poesio, "Named Entity Recognition as Dependency Parsing," in *Proceedings of the 58th Annual Meeting of the Association for Computational Linguistics*, 2020, pp. 6470–6476.

[7]   M. Hagiwara, "Multi-Task Transfer Learning for Fine-Grained Named Entity Recognition," in *Proceedings of the Twelfth Text Analysis Conference*, 2019.

[8]   A. Vaswani *et al.*, "Attention is all you need," in *Advances in Neural Information Processing Systems*, 2017, vol. 2017-Decem, pp. 5999–6009.

[9]   J. Devlin, M.-W. Chang, K. Lee, K. T. Google, and A. I. Language, "BERT: Pre-training of Deep Bidirectional Transformers for Language Understanding," in *Proceedings of NAACL-HLT 2019*, 2019, pp. 4171–4186.

[10] E. Alsentzer *et al.*, "Publicly Available Clinical BERT Embeddings," in *Proceedings of the 2nd Clinical Natural Language Processing Workshop*, 2019, pp. 72–78.

[11] A. E. W. Johnson *et al.*, "MIMIC-III, a freely accessible critical care database," *Sci. Data*, vol. 3, p. 160035, May 2016.

[12] Ö. Uzuner, B. R. South, S. Shen, and S. L. DuVall, "2010 i2b2/VA challenge on concepts, assertions, and relations in clinical text," *J. Am. Med. Informatics Assoc.*, vol. 18, no. 5, pp. 552–556, Sep. 2011.

[13] W. W. Chapman, W. Bridewell, P. Hanbury, G. F. Cooper, and B. G. Buchanan, "A simple algorithm for identifying negated findings and diseases in discharge summaries," *J. Biomed. Inform.*, vol. 34, no. 5, pp. 301–310, Oct. 2001.

[14] O. Bodenreider, "The Unified Medical Language System (UMLS): integrating biomedical terminology.," *Nucleic Acids Res.*, vol. 32, no. Database issue, pp. D267-70, Jan. 2004.

[15] H. Harkema, J. N. Dowling, T. Thornblade, and W. W. Chapman, "ConText: An algorithm for determining negation, experiencer, and temporal status from clinical reports," *J. Biomed. Inform.*, vol. 42, no. 5, pp. 839–851, Oct. 2009.

[16] S. Mehrabi *et al.*, "DEEPEN: A negation detection system for clinical text incorporating dependency relation into NegEx," *J. Biomed. Inform.*, vol. 54, pp. 213–219, Apr. 2015.

[17] C. Clark *et al.*, "MITRE system for clinical assertion status classification," *J. Am. Med. Informatics Assoc.*, vol. 18, no. 5, pp. 563–567, Sep. 2011.

[18] A. Khandelwal and B. K. Britto, "Multitask Learning of Negation and Speculation using Transformers," in *Proceedings of the 11th International Workshop on Health Text Mining and Information Analysis*, 2020, pp. 79–87.

[19] P. Bhatia and M. Khalilia, "Joint Entity Extraction and Assertion Detection for Clinical Text," in *Proceedings of the 57th Annual Meeting of the Association for Computational Linguistics*, 2019, pp. 954–959.

[20] T. Liu, J.-G. Yao, and C.-Y. Lin, "Towards Improving Neural Named Entity Recognition with Gazetteers."

[21] N. Nath, S.-H. Lee, M. McDonnell, and I. Lee, "The quest for better clinical word vectors: Ontology based and lexical vector augmentation versus clinical contextual embeddings," *Comput. Biol. Med.*, vol. 134, p. 104433, Jul. 2021.

[22] L. Ratinov and D. Roth, "Design challenges and misconceptions in named entity recognition," in *CoNLL 2009 - Proceedings of the Thirteenth Conference on Computational Natural Language Learning*, 2009, pp. 147–155.

[23] R. J. Williams and D. Zipser, "A Learning Algorithm for Continually Running Fully Recurrent Neural Networks," *Neural Comput.*, vol. 1, no. 2, pp. 270–280, Jun. 1989.

[24] B. De Bruijn, C. Cherry, S. Kiritchenko, and J. Martin, "NRC at i2b2: one challenge, three practical tasks, nine statistical systems, hundreds of clinical records, millions of useful

features."

[25]  S. Wu *et al.*, "Negation's not solved: Generalizability versus optimizability in clinical natural language processing," *PLoS One*, vol. 9, no. 11, Nov. 2014.

[26]  Z. Liu *et al.*, "Entity recognition from clinical texts via recurrent neural network," *BMC Med. Inform. Decis. Mak.*, vol. 17, no. S2, p. 67, Jul. 2017.

[27]  W. Boag, E. Sergeeva, S. Kulshreshtha, P. Szolovits, A. Rumshisky, and T. Naumann, "CliNER 2.0: Accessible and Accurate Clinical Concept Extraction," 2018.

[28]  J. Pennington, R. Socher, and C. Manning, "Glove: Global Vectors for Word Representation," in *Proceedings of the 2014 Conference on Empirical Methods in Natural Language Processing (EMNLP)*, 2014, pp. 1532–1543.

[29]  Z. Huang, W. Xu, and K. Yu, "Bidirectional LSTM-CRF Models for Sequence Tagging," Aug. 2015.

[30]  S. Hochreiter and J. Urgen Schmidhuber, "LONG SHORT-TERM MEMORY," *Neural Comput.*, vol. 9, no. 8, pp. 1735–1780, 1997.

[31]  A. J. Viterbi, "Error Bounds for Convolutional Codes and an Asymptotically Optimum Decoding Algorithm," *IEEE Trans. Inf. Theory*, vol. 13, no. 2, pp. 260–269, 1967.

[32]  Z. Qian, P. Li, Q. Zhu, G. Zhou, Z. Luo, and W. Luo, "Speculation and negation scope detection via convolutional neural networks," in *EMNLP 2016 - Conference on Empirical Methods in Natural Language Processing, Proceedings*, 2016, pp. 815–825.

[33]  P. Banerjee, K. Kumar Pal, M. Devarakonda, and C. Baral, "Biomedical Named Entity Recognition via Knowledge Guidance and Question Answering," *ACM Trans. Comput. Healthc.*, vol. 2, 2021.

[34]  Y. Xu, Y. Wang, T. Liu, J. Tsujii, and E. I. C. Chang, "An end-to-end system to identify temporal relation in discharge summaries: 2012 i2b2 challenge," *J. Am. Med. Informatics Assoc.*, vol. 20, no. 5, pp. 849–858, 2013.

[35]  M. Gardner *et al.*, "AllenNLP: A deep semantic natural language processing platform," *arXiv*. pp. 1–6, 2018.

[36]  M. S. Akhtar, T. Garg, and A. Ekbal, "Multi-task learning for aspect term extraction and aspect sentiment classification," *Neurocomputing*, vol. 398, pp. 247–256, Jul. 2020.

[37]  Y. Wang, M. Huang, A. Sun, and X. Zhu, "Aspect-level sentiment analysis using AS-capsules," *Web Conf. 2019 - Proc. World Wide Web Conf. WWW 2019*, pp. 2033–2044, May 2019.

[38]  Y. Xiong, Y. Huang, Q. Chen, X. Wang, Y. Ni, and B. Tang, "A Joint Model for Medical Named Entity Recognition and Normalization," 2020.

[39]  A. Miranda-Escalada, E. Farré, and M. Krallinger, "Named Entity Recognition, Concept Normalization and Clinical Coding: Overview of the Cantemist Track for Cancer Text Mining in Spanish, Corpus, Guidelines, Methods and Results," 2020.

[40] "Why are precision, recall and F1 score equal when using micro averaging in a multi-class problem? – Simon's blog." [Online]. Available: https://simonhessner.de/why-are-precision-recall-and-f1-score-equal-when-using-micro-averaging-in-a-multi-class-problem/. [Accessed: 03-Feb-2021].